\documentclass[10.5pt,letterpaper,fleqn, twocolumn]{article}

\usepackage{graphicx}
\usepackage{makeidx}
\usepackage{multicol}
\usepackage{crop}
\usepackage{amstext,amsthm}
\usepackage{amsmath,amssymb}
\usepackage{bm}
\usepackage{natbib}
\usepackage{multirow}
\usepackage{hhline}

\usepackage{multicol}
\usepackage{pslatex}
\usepackage[strict]{changepage}
\newcommand{\be}{\begin{equation}}
\newcommand{\ee}{\end{equation}}
\newcommand{\bea}{\begin{eqnarray}}
\newcommand{\eea}{\end{eqnarray}}
\newcommand{\bne}{\begin{equation*}}
\newcommand{\ene}{\end{equation*}}
\newcommand{\bi}{\begin{itemize}}
\newcommand{\ei}{\end{itemize}}

\newcommand{\bbm}{\begin{bmatrix}}
\newcommand{\ebm}{\end{bmatrix}}

\newcommand{\mr}{\mathrm}

\usepackage[top=0.75in, bottom=0.5in, left=0.83in, right=0.83in]{geometry}
\usepackage[hang,splitrule]{footmisc}





\usepackage{titlesec}		% For titleformat and titlespacing
\usepackage{subcaption}		% For subfigures
\usepackage{indentfirst}	% Indent at start of section/subsection

\newenvironment{myabstract}
{
	\vspace{0.2in}
	\parindent=0cm \textit{Abstract}
	\parindent=0.5cm \hangindent=0.5cm \linebreak
}

\newenvironment{mykeywords}
{
	\vspace{0.2in}
	\parindent=0cm \textit{Keywords}
	
	\parindent=0.5cm
}

\titleformat{\section}
{\normalfont\bfseries}
{}{0.0pt}{\parindent=0cm}[\parindent=0.5cm]

\titlespacing*{\section}
{0pt}{1.5\baselineskip}{0.5\baselineskip}

\titleformat{\subsection}
{\normalfont\itshape}
{}{0.0pt}{\parindent=0cm}[\parindent=0.5cm]

\titlespacing*{\subsection}
{0pt}{1.0\baselineskip}{0.5\baselineskip}

\begin{document}

\twocolumn[{%
	
% Title
\phantom\\
\vspace{0.5in}
\begin{center}
\Large{\textbf{Physics-Informed Neural Network with Transfer Learning for State Estimation in Lithium-Ion Batteries using the Single Particle Model with Electrolyte}}\\
\end{center}
\vspace{0.2in}

% Authors and affiliations
\begin{center}
Gift Modekwe $^{\mr{a}}$ and Qiugang Lu $^{\mr{a,}}$\footnotemark\\

\vspace{0.10in}

$^{\mr{a}}$ Department of Chemical Engineering, Texas Tech University, Lubbock, TX 79409, USA

\end{center}

% Abstract
\begin{myabstract}
Physics-informed neural networks (PINNs) have emerged as a powerful tool for solving nonlinear partial differential equations (PDEs), including battery electrochemical models. They typically en-force conservation laws within the loss function to ensure physically consistent solutions. Tradi-tional numerical methods such as finite difference, finite volume, and finite element techniques, re-ly on discretization and can be computationally expensive for nonlinear systems. To address this challenge, PINNs offer improved scalability, particularly for reduced-order models like the single particle model with electrolyte (SPMe). The SPMe describes lithium-ion battery dynamics through coupled diffusion, transport, reaction kinetics, and voltage equations. Despite these advantages, training SPMe-based PINNs from scratch for different battery chemistries or operating conditions is demanding and often leads to slow convergence. To overcome this limitation, this work introduces a transfer learning framework for SPMe-PINNs. The model is first pretrained to learn general elec-trochemical dynamics and then adapted to a target battery by transferring weights, freezing se-lected layers, and fine tuning the remaining parameters, including estimating key electrochemical variables. Validation using PyBaMM demonstrates accurate voltage prediction, indicating that the proposed approach preserves electrochemical consistency while reducing training time and ena-bling efficient generalization across batteries.
\end{myabstract}

% Keywords
\begin{mykeywords}
Lithium-ion battery, Physics-informed neural networks, Transfer learning, Single particle model with electrolyte.
\end{mykeywords}

% Make a little space before starting body
\vspace{0.2in}
}]

% Footnote text
\footnotetext[1]{\parindent=0cm \small{Corresponding author: Q. Lu (E-mail: {\tt\small jay.lu@ttu.edu}).}}

%-----------------------------------------------------------------------------------------------------------------------------------------------------------
%
%
%-----------------------------------------------------------------------------------------------------------------------------------------------------------

%\begin{multicols}{2}

% Include txt files with sections
%\input{tex/introduction}

\section{1. Introduction}
Lithium-ion batteries have emerged as a dominant technology in modern energy storage, owing to their superior energy density, extended cycle life, and high operational efficiency \citep{khan2023design}. Their versatility has led to widespread use in portable electronics, electric vehicles, and grid-scale energy storage systems. As the demand for reliable energy storage grows, accurate battery modeling and monitoring have become essential for ensuring operational safety, optimizing performance, and extending battery lifespan. In particular, accurate estimation of internal battery states and prediction of battery behavior are critical for effective battery management \citep{ekberg2025state}. 

Significant research efforts have focused on robust modeling of lithium-ion batteries. Notably, both physics-based and data-driven modeling paradigms have been extensively investigated in the literature. Physics-based models, such as the Doyle–Fuller–Newman (DFN) model \citep{doyle1993modeling} and its reduced-order variants, like the single particle model (SPM) \citep{santhanagopalan2006review} and the SPM with electrolyte (SPMe) \citep{moura2016battery}, are grounded in fundamental electrochemical principles. These models characterize lithium-ion transport, charge conservation, and reaction kinetics through systems of coupled PDEs. Thus, they offer strong physical interpretability and predictive consistency across a wide range of operating conditions. Despite these advantages, their practical deployment is often hindered by the need for detailed knowledge of material-specific parameters, as well as significant computational demands that limit their suitability for real-time applications.

On the other hand, data-driven models leverage machine learning techniques to learn intricate relationships directly from data without requiring explicit knowledge of underlying physics. Specifically, deep learning frameworks have demonstrated strong capability in capturing nonlinear battery behaviors and degradation patterns \citep{alharbi2025lithium}. However, these models often suffer from limited interpretability, high data requirements, and reduced generalization to unseen operating conditions \citep{liu2022review}.

To address these limitations, PINNs have emerged as a promising hybrid framework that bridges the gap between physics-based and data-driven modeling. They incorporate governing physical equations directly into the neural network training process in order to enforce physical consistency, while retaining the flexibility of data-driven learning \citep{raissi2019physics}. This paradigm has attracted considerable attention in battery modeling and state estimation. For instance, \cite{singh2023hybrid}  developed a PINN framework for lithium-ion battery state estimation, demonstrating improved accuracy over purely data-driven approaches by embedding electrochemical constraints into the loss function. Similarly, \cite{nascimento2021hybrid} proposed a hybrid PINN framework for lithium-ion battery health prognostics, demonstrating reliable prognosis performance even with limited data. Building upon these advancements, \cite{xue2023enhanced} introduced the PINN SPM, which utilizes a gated recurrent unit to solve electrolyte diffusion equations, effectively correcting voltage predictions of traditional SPM at high C-rates. Additionally, \cite{mendez2024physics} demonstrated that a purely physics-driven PINN can solve the SPM across broad parameter ranges without any reliance on labeled experimental or simulation data. Despite these advances, a critical and largely unresolved challenge remains: existing PINN-based approaches are predominantly trained on individual battery basis, limiting their transferability across batteries with differing chemistries, aging histories, and operating conditions. 

In this study, we focus on improving the scalability and generalization of PINNs for lithium-ion battery modeling by integrating transfer learning within an SPMe-based PINN framework. The proposed approach leverages knowledge learned from a source battery to efficiently adapt to a target battery while preserving electrochemical consistency. The main contributions of this work are as follows:

\begin{itemize}
	\item To the best of our knowledge, this is the first work to integrate transfer learning within a phys-ics-informed SPMe framework for lithium-ion battery modeling.
	\item A PINN based on the SPMe is developed to cap-ture coupled solid-phase diffusion, electrolyte transport, and electrochemical reaction kinetics within a unified learning framework. 	
	\item A transfer learning strategy is introduced, where a pretrained SPMe-PINN is adapted to a target battery to capture and adapt to battery-specific characteristics.	
	\item Key electrochemical variables are incorporated as learnable parameters during fine-tuning, ena-bling adaptation to different chemistries and op-erating conditions with simultaneous parameter estimation.	
\end{itemize}

The remainder of this paper is organized as follows. Section \ref{SPMe} introduces the fundamentals of the SPMe model. Section \ref{TL_SPMe} presents the proposed framework. Section \ref{results} provides the results and performance evaluation. Finally, conclusions are drawn in Section \ref{conclusion}.

\section{2. Fundamentals of the SPMe}
\label{SPMe}

\begin{figure}[tbh]
	\centering
	\includegraphics[width =0.4\textwidth]{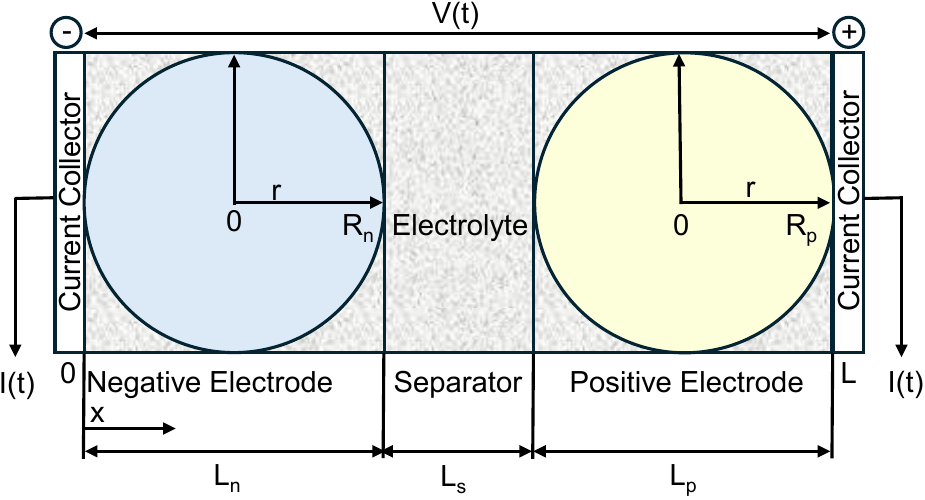}
	\caption{Schematics of the SPMe model.}
	\label{schematics}
\end{figure}

The SPMe is an electrochemical model used to describe the behavior of lithium-ion batteries, extending the classic SPM model by incorporating electrolyte dynamics. In SPM, each electrode is represented as a single spherical particle, simplifying the complex porous electrode structure while capturing essential solid-phase lithium diffusion and intercalation kinetics. The SPMe enhances this by adding electrolyte concentration and potential variations across the cell to account for ionic transport through the separator and electrolyte resistance as in Fig. \ref{schematics}. These effects are typically more significant at higher C-rates \citep{mehta2021improved}.

\subsection{2.1 Solid-Phase Diffusion}
In the SPMe, lithium-ion diffusion within the active material particles of positive ($p$) and negative $(n)$ electrodes is governed by Fick’s second law in spherical coordinates:
\begin{equation}
	\frac{\partial c_{s,k}}{\partial t}
	=
	\frac{D_{s,k}}{r_k^2}
	\frac{\partial}{\partial r_k}
	\left(
	r^2
	\frac{\partial c_{s,k}}{\partial r_k}
	\right),
	\qquad
	k \in \{n,p\},
\end{equation}
where $c_{s,k}(r,t)$ denotes the solid-phase lithium concentration, $D_{s,k}$ is the solid diffusivity, and $r$ is the radial coordinate inside the spherical particle.

The corresponding boundary conditions are
\begin{equation}
	\left.
	\frac{\partial c_{s,k}}{\partial r_k}
	\right|_{r_k=0}=0, \qquad
	-D_{s,k}
\left.
\frac{\partial c_{s,k}}{\partial r_k}
\right|_{r_k=R_k}=\frac{j_k}{F},	
\end{equation}
where $R_k$ is the particle radius, $j_k$ is the interfacial reaction flux, and $F$ is Faraday’s constant.

\subsection{2.2 Electrolyte Dynamics}

The electrolyte concentration dynamics are described across the negative electrode, separator, and positive electrode domains by
\begin{equation}
	\epsilon_{e,k}
	\frac{\partial c_{e,k}}{\partial t}
	=
	\frac{\partial}{\partial x}
	\left(
	D_{e,k}^{\mathrm{eff}}
	\frac{\partial c_{e,k}}{\partial x}
	\right)
	+
	\begin{cases}
		\dfrac{1-t_+}{F L_k} I(t), & k \in \{n,p\}, \\[8pt]
		0, & k=s,
	\end{cases}
\end{equation}
where $\epsilon_{e,k}$ is the electrolyte volume fraction, $c_{e,k}$ is the electrolyte concentration, $D_{e,k}^{\mathrm{eff}}$ is the effective electrolyte diffusivity, $t_+$ is the lithium-ion transference number, $L_k$ is the thickness of region $k$, and $I(t)$ is the applied current density. The source term is zero in the separator region $(k=s)$ because no electrochemical reactions occur within the separator.
The effective electrolyte diffusivity is corrected using the Bruggeman relation:
\begin{equation}
	D_e^{\mathrm{eff}}
	=
	D_e \epsilon_e^{b},
\end{equation}
where $b$ is the Bruggeman coefficient. Zero-flux boundary conditions are imposed at the current collectors
\begin{equation}
	\left.
	\frac{\partial c_e}{\partial x}
	\right|_{x=0}
	=
	0,
	\qquad
	\left.
	\frac{\partial c_e}{\partial x}
	\right|_{x=L}
	=
	0,
\end{equation}
where $L$ is the total cell thickness. Flux and concentration continuity are also assumed at the boundaries.

\subsection{2.3 Electrochemical Kinetics}

The electrochemical reaction kinetics are modeled using the Butler--Volmer relation. The exchange current density is expressed as
\begin{equation}
	i_{0,k}
	=
	k_k
	c_e^{\alpha}
	c_{s,k}^{\alpha}
	\left(
	c_{s,k}^{\max}
	-
	c_{s,k}
	\right)^{\alpha},
\end{equation}
where $k_k$ is the reaction-rate constant, $c_{s,k}^{\max}$ is the maximum solid concentration, and $\alpha$ is the charge-transfer coefficient.

The reaction overpotential is given by
\begin{equation}
	\eta_k
	=
	\frac{2RT}{F}
	\sinh^{-1}
	\left(
	\frac{j_k}{2i_{0,k}}
	\right),
\end{equation}
where $R$ is the universal gas constant and $T$ is temperature. The open-circuit potentials of the electrodes are represented as nonlinear functions of stoichiometry
\begin{equation}
	U_n = U_n(x_n),
	\qquad
	U_p = U_p(x_p),
\end{equation}
where $x_n$ and $x_p$ denote the negative and positive electrode stoichiometries.

\subsection{2.4 Terminal Voltage}

The measurable terminal voltage at the current collectors is obtained by combining the equilibrium potentials, reaction overpotentials, electrolyte potential drop, and solid-phase ohmic losses: 
\begin{equation}
	V(t)
	=
	(U_p-U_n)
	+
	(\eta_p-\eta_n)
	+
	\Delta \phi_{\mathrm{elec}}
	+
	\Delta \phi_{\mathrm{solid}},
\end{equation}
where $U_p$ and $U_n$ denote the positive and negative electrode open-circuit potentials, $\eta_p$ and $\eta_n$ represent the activation overpotentials obtained from Butler--Volmer kinetics, $\Delta \phi_e$ is the electrolyte potential drop, and $\Delta \phi_s$ is the solid-phase ohmic drop. The electrolyte potential drop \citep{moura2016battery} is expressed as:
\begin{equation}
	\Delta \phi_{\mathrm{elec}}
	=
	\frac{(L_n+2L_s+L_p)}{2\kappa}I(t)
	+
	\frac{2RT}{F}
	k_f(1-t_+)
	\ln
	\left(
	\frac{c_{e}^0}{c_{e}^L}
	\right), 
\end{equation}
and the solid-phase ohmic drop is
\begin{equation}
	\Delta \phi_{\mathrm{solid}}
	=
	-
	\frac{I(t)}{3}
	\left(
	\frac{L_p}{\sigma_p}
	+
	\frac{L_n}{\sigma_n}
	\right),
\end{equation}
with $\kappa$ denoting electrolyte conductivity, $\sigma_n$ and $\sigma_p$ as the electrode conductivities \citep{marquis2019asymptotic} and $k_f$ as the electrolyte thermodynamic factor that accounts for non-ideal electrolyte behavior.  
 
\section{3. Proposed Transfer Learning-based SPMe-PINN}
\label{TL_SPMe}
\begin{figure}[tbh]
	\centering
	\includegraphics[width =0.4\textwidth]{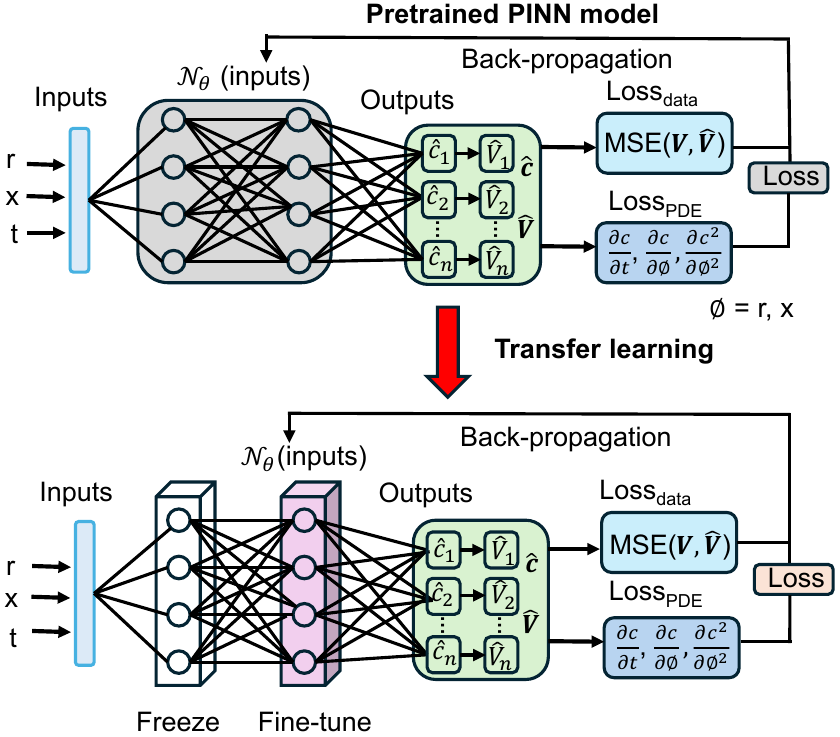}
	\caption{Illustration of the proposed transfer learning framework for the SPMe-PINN model.}
	\label{transferrable_PINN}
\end{figure}

In this section, the proposed transfer learning framework for the SPMe-based physics-informed neural network (SPMe-PINN) is presented. The objective is to efficiently transfer electrochemical knowledge learned from a source battery to a target battery while preserving the electrochemical consistency imposed by the governing equations. The framework consists of two major stages (see Fig. \ref{transferrable_PINN}): (i) pretraining on a source battery and (ii) fine-tuning on a target battery.

\subsection{3.1 Pretraining on Source Battery}

In the pretraining stage, a PINN is trained using electrochemical data generated from a source battery. The PINN takes normalized temporal and spatial coordinates as inputs and predicts the electrochemical states (concentrations) of the battery at the surface, including the solid-phase lithium concentrations in the negative and positive particles and the electrolyte concentration at the boundaries.

The normalized variables are defined as
\begin{equation}
	\hat{t}=\frac{t}{t_{\max}}, \qquad
	\hat{r}=\frac{r}{R_k}, \qquad
	\hat{x}=\frac{x}{L},
\end{equation}
while the normalized concentrations are expressed as
\begin{equation}
	\hat{c}_{s,k}=\frac{c_{s,k}}{c_{s,k}^{\max}}, \qquad
	\hat{c}_{e}=\frac{c_e}{c_{e,0}}.
\end{equation}

Three neural networks are employed to approximate the electrochemical states for the negative electrode, positive electrode and electrolyte. 
%\begin{equation}
%	(\hat{t},\hat{r}) \rightarrow \hat{c}_{s,n}, \qquad
%		(\hat{t},\hat{r}) \rightarrow \hat{c}_{s,p}, \qquad
%			(\hat{t},\hat{x}) \rightarrow \hat{c}_{e}.
%\end{equation}
The PDE residuals are evaluated at a set of collocation points sampled within the normalized computational domains. The residual loss is written as:
%\begin{equation}
%	\frac{\partial \hat{c}_{s,i}}{\partial \hat{t}}
%	-
%	\frac{D_{s,k}t_{\max}}{R_k^2}
%	\frac{1}{\hat{r}^2}
%	\frac{\partial}{\partial \hat{r}}
%	\left(
%	\hat{r}^2
%	\frac{\partial \hat{c}_{s,k}}{\partial \hat{r}}
%	\right)
%	=0,
%\end{equation}
%
%while the electrolyte diffusion equation is given by
%\begin{equation}
%	\varepsilon_e
%	\frac{\partial \hat{c}_e}{\partial \hat{t}}
%	-
%	\frac{t_{\max}}{L^2}
%	D_e^{eff}
%	\frac{\partial^2 \hat{c}_e}{\partial \hat{x}^2}
%	-
%	\frac{t_{\max}}{Lc_{e,0}}S
%	=0,
%\end{equation}
%where the source term is defined as
%\begin{equation}
%	S=\frac{1-t_+^0}{F}j.
%\end{equation}
\begin{equation}
	\mathcal{L}_{pde}
	=
	\frac{1}{N_{pde}}
	\sum_{j=1}^{N_{pde}}
	\left(
	\|\mathcal{R}_{s,n}^{(j)}\|^2
	+
	\|\mathcal{R}_{s,p}^{(j)}\|^2
	+
	\|\mathcal{R}_{e}^{(j)}\|^2
	\right),
\end{equation}
where $\mathcal{R}_{s,n}$, $\mathcal{R}_{s,p}$, and $\mathcal{R}_{e}$ denote the residuals of the negative solid-phase, positive solid-phase, and electrolyte PDEs, respectively. In addition to enforcing the governing equations, the normalized PINN formulation also satisfies the corresponding boundary and initial conditions for the solid-phase and electrolyte concentration dynamics at selected points.

%The solid-phase diffusion equations satisfy symmetry and flux boundary conditions given by
%
%\begin{equation}
%	\left.
%	\frac{\partial \hat{c}_{s,i}}{\partial \hat{r}}
%	\right|_{\hat{r}=0}
%	=0, \qquad
%		\left.
%	\frac{\partial \hat{c}_{s,i}}{\partial \hat{r}}
%	\right|_{\hat{r}=1}
%	=	-
%	\frac{R_i}{D_{s,i}c_{s,i}^{max}}
%	\frac{j_i}{F}.
%\end{equation}
%
%Similarly, the electrolyte concentration satisfies zero-flux boundary conditions
%
%\begin{equation}
%	\left.
%	\frac{\partial \hat{c}_e}{\partial \hat{x}}
%	\right|_{\hat{x}=0}
%	=0,
%	\qquad
%	\left.
%	\frac{\partial \hat{c}_e}{\partial \hat{x}}
%	\right|_{\hat{x}=1}
%	=0.
%\end{equation}
%The initial conditions are enforced as
%\begin{equation}
%	\hat{c}_{s,n}(0,\hat{r})=\hat{c}_{s,n,0}, \qquad
%	\hat{c}_{s,p}(0,\hat{r})=\hat{c}_{s,p,0}, \qquad
%	\hat{c}_{e}(0,\hat{x})=1.	
%\end{equation}
After PINN predicts solid-phase and electrolyte concentrations, these states are used to compute the terminal voltage through the SPMe voltage formulation. The predicted surface concentrations are used to evaluate the open-circuit potentials and exchange current densities, while the electrolyte concentration accounts for electrolyte polarization. 
%The terminal voltage is expressed as
%\begin{equation}
%	V_{PINN}
%	=
%	\left(U_p-U_n\right)
%	+
%	\left(\eta_p-\eta_n\right)
%	+
%	\Delta \phi_e
%	+
%	\Delta \phi_s,
%\end{equation}

The PINN is trained by minimizing a composite loss function consisting of PDE residual, boundary condition, initial condition, and voltage losses. The optimization problem during pretraining is formulated as:
\begin{equation}
	\Theta_s^{*} =
	\arg\min_{\Theta}
	\left(
	w_{pde}\mathcal{L}_{pde}
	+
	w_{bc}\mathcal{L}_{bc}
	+
	w_{ic}\mathcal{L}_{ic}
	+
	w_V\mathcal{L}_{V}
	\right),
\end{equation}
where: 
\begin{equation}
	\mathcal{L}_{bc} = \frac{1}{N_{bc}}\sum_{j=1}^{N_{bc}} \left\| BC_{pred}^{(j)} - BC_{true}^{(j)} \right\|^2,
\end{equation}
\vspace{-0.1cm}
\begin{equation}
	\mathcal{L}_{ic} = \frac{1}{N_{ic}}\sum_{j=1}^{N_{ic}} \left\| IC_{pred}^{(j)} - IC_{true}^{(j)} \right\|^2,
\end{equation}
\vspace{-0.1cm}
\begin{equation}
	\mathcal{L}_{V}
	=
	\frac{1}{N_V}
	\sum_{j=1}^{N_V}
	\left\|
	V_{PINN}^{(j)}
	-
	V_{ref}^{(j)}
	\right\|^2,
\end{equation}
$V_{ref}$ denotes the reference terminal voltage obtained from the source battery data and $\Theta_s$ denotes the trainable parameters of the pretrained model, including the neural-network weights, biases, and learnable electrochemical kinetic parameters. Since the voltage computation remains fully differentiable, the voltage loss is backpropagated through the SPMe voltage equations and subsequently through the predicted concentration fields, enabling the network parameters and learnable electrochemical parameters to be updated simultaneously during training. Through this process, the PINN learns generalized electrochemical dynamics that can be transferred across different battery systems.

\subsection{3.2 Fine-Tuning on Target Battery}
In the second stage, the pretrained SPMe-PINN is adapted to a target battery with potentially different chemistry, operating conditions, or material properties. The pretrained parameters are first transferred to initialize the target model according to
\begin{equation}
	\Theta_t \leftarrow \Theta_s,
\end{equation}
after which the network is further optimized using target battery data by minimizing the fine-tuning loss
\begin{equation}
	\mathcal{L}_{\text{fine-tune}} = \mathcal{L}_{\text{PDE}}^{(t)} + \lambda\,\mathcal{L}_{\text{data}}^{(t)},
\end{equation}
where $\mathcal{L}_{\text{data}}^{(t)} = \text{MSE}(\mathbf{V}, \hat{\mathbf{V}})$ is the data fidelity term computed on target battery measurements, $\mathcal{L}_{\text{PDE}}^{(t)}$ enforces the SPMe governing equations on the target domain, and $\lambda$ is a weighting factor. To preserve the shared electrochemical knowledge learned during pretraining, selected layers of the pretrained network are frozen during fine-tuning. 

During fine-tuning, parameters associated with target-specific electrochemical properties are updated, while the shared electrochemical dynamics encoded within the pretrained model are preserved. In this work, the fine-tuned parameters are the solid-phase diffusivities of the negative and positive electrodes.

\subsection{3.4 Computational Efficiency and Model Benefits}
Compared with training from scratch, the proposed transfer learning can significantly reduce training time and enhance convergence stability. Additionally, by leveraging pretrained electrochemical knowledge, the model achieves improved prediction accuracy under different data scenarios. 

\section{4. Results and Discussion}
\label{results}
In this section, the performance of the proposed framework is evaluated through a series of case studies designed to assess its accuracy, robustness, and generalization capability. The analysis begins with validation on a source battery using the \cite{chen2020development} parameter set (hereafter referred to as B1) to establish the baseline performance of the model. The PINN architecture consists of four hidden layers with 64 neurons per layer. This is followed by transfer learning studies on different battery datasets, incorporating variations in cell configuration, operating conditions, and chemistry. Specifically, the \cite{ecker2015parameterization} dataset (hereafter referred to as B2) is used to examine cross-domain transfer within similar chemistry, while the \cite{prada2013simplified} dataset (hereafter referred to as B3) is employed to evaluate cross-chemistry generalization. During fine-tuning, the first two layers of the pretrained network are frozen to preserve the learned electrochemical representations while adapting the remaining parameters to the target battery domain. The results include comparisons of terminal voltage and internal state predictions against reference solutions obtained from PyBaMM.

\subsection{4.1 Source Model Validation}

The B1 dataset is based on the LG M50 graphite/NMC lithium-ion cylindrical cell. We consider a constant 1C discharge condition. The objective of this experiment is to verify that the proposed PINN architecture can accurately reproduce the electrochemical dynamics of the SPMe model prior to transfer learning.

\begin{figure}[tbh]
	\centering
	\includegraphics[width =0.4\textwidth]{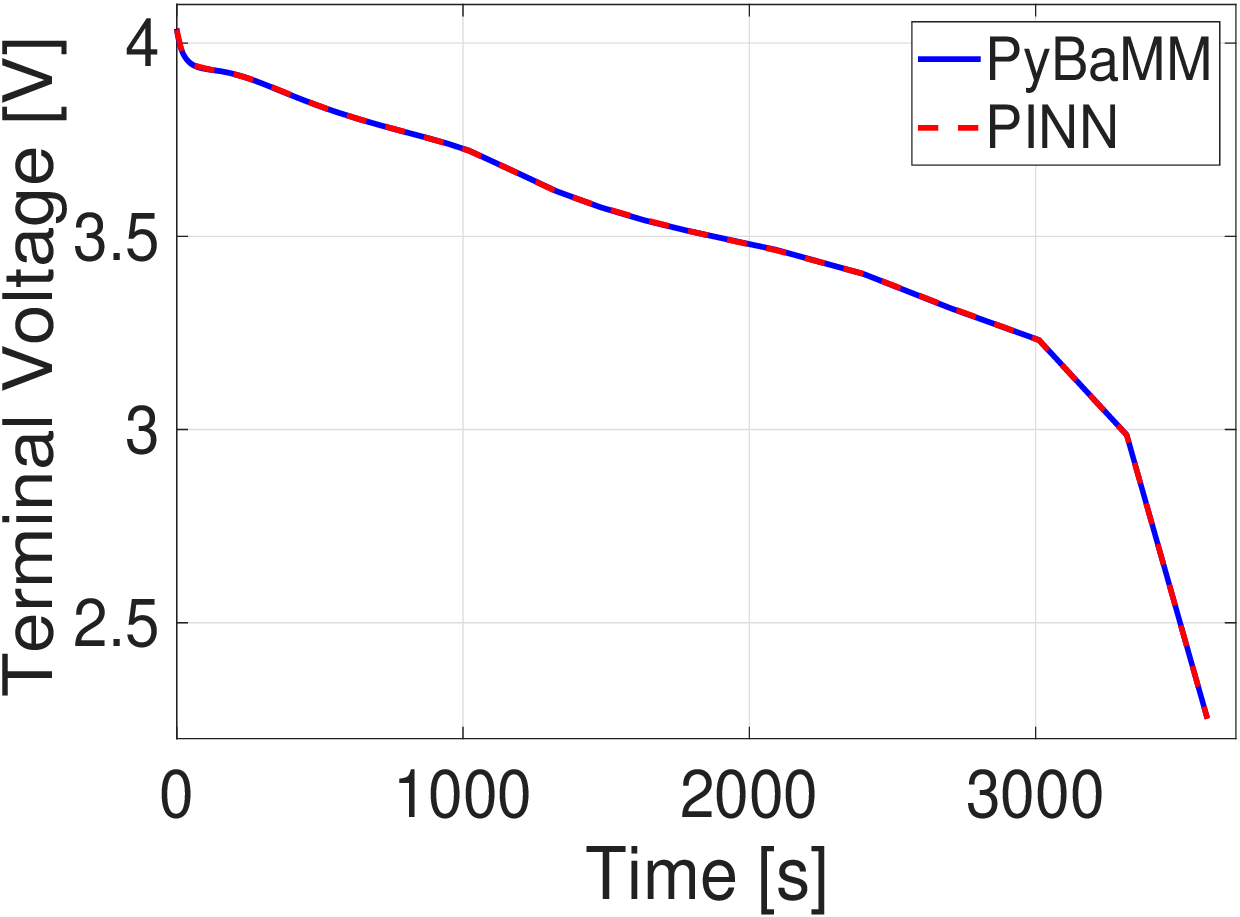}
	\caption{Comparison of terminal voltage predictions obtained from the PyBaMM SPMe model and the SPMe-PINN framework for B1.}
	\label{source_voltage}
\end{figure}

The terminal voltage predicted by the SPMe-PINN is compared against the reference SPMe solution obtained from PyBaMM. As shown in Fig. \ref{source_voltage}, the proposed model closely tracks the reference voltage profile across the entire discharge window. The prediction error remains consistently low (a root mean square error of 8.1e$^{-4}$), indicating that the model successfully captures both transient and steady-state electrochemical behavior. This agreement demonstrates that the PINN is able to learn the dynamics governing the system, thereby providing a reliable pretrained model for subsequent transfer learning tasks.

\subsection{4.2 Cross-Domain Transfer}
\begin{figure}[h]
	\centering
	% Top row
	\begin{subfigure}[b]{0.2\textwidth}
		\centering
		\includegraphics[height=3cm,keepaspectratio]{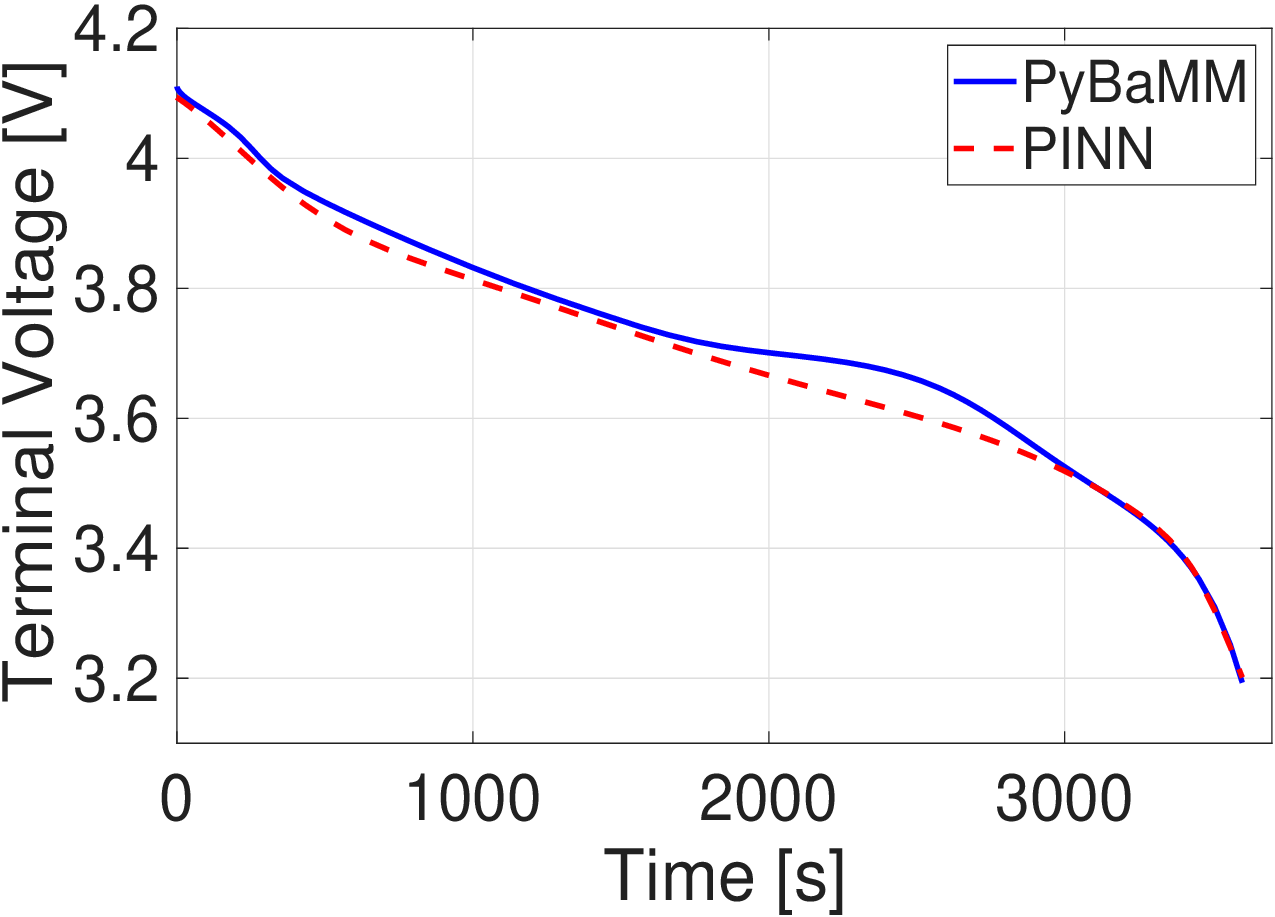}
		\caption{}
		\label{ecker_voltage}
	\end{subfigure}
	\hspace{0.04\textwidth}
	\begin{subfigure}[b]{0.2\textwidth}
		\centering
		\includegraphics[height=3cm,keepaspectratio]{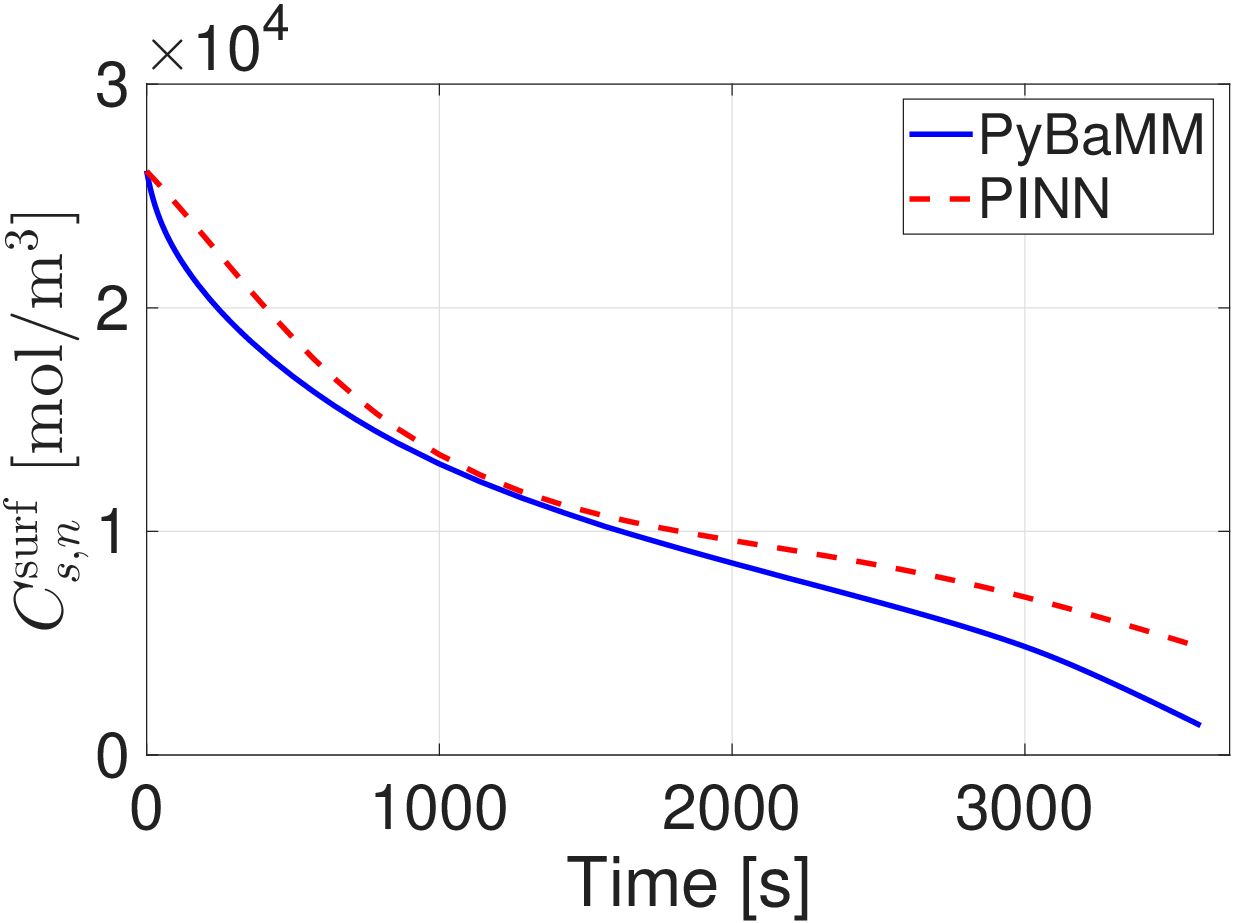}
		\caption{}
		\label{ecker_concentration_negative}
	\end{subfigure}
	
	%\hspace{0.005\textwidth}
	\par\vspace{0.2cm}
	% Bottom row
	\begin{subfigure}[b]{0.2\textwidth}
		\centering
		\includegraphics[height=3cm,keepaspectratio]{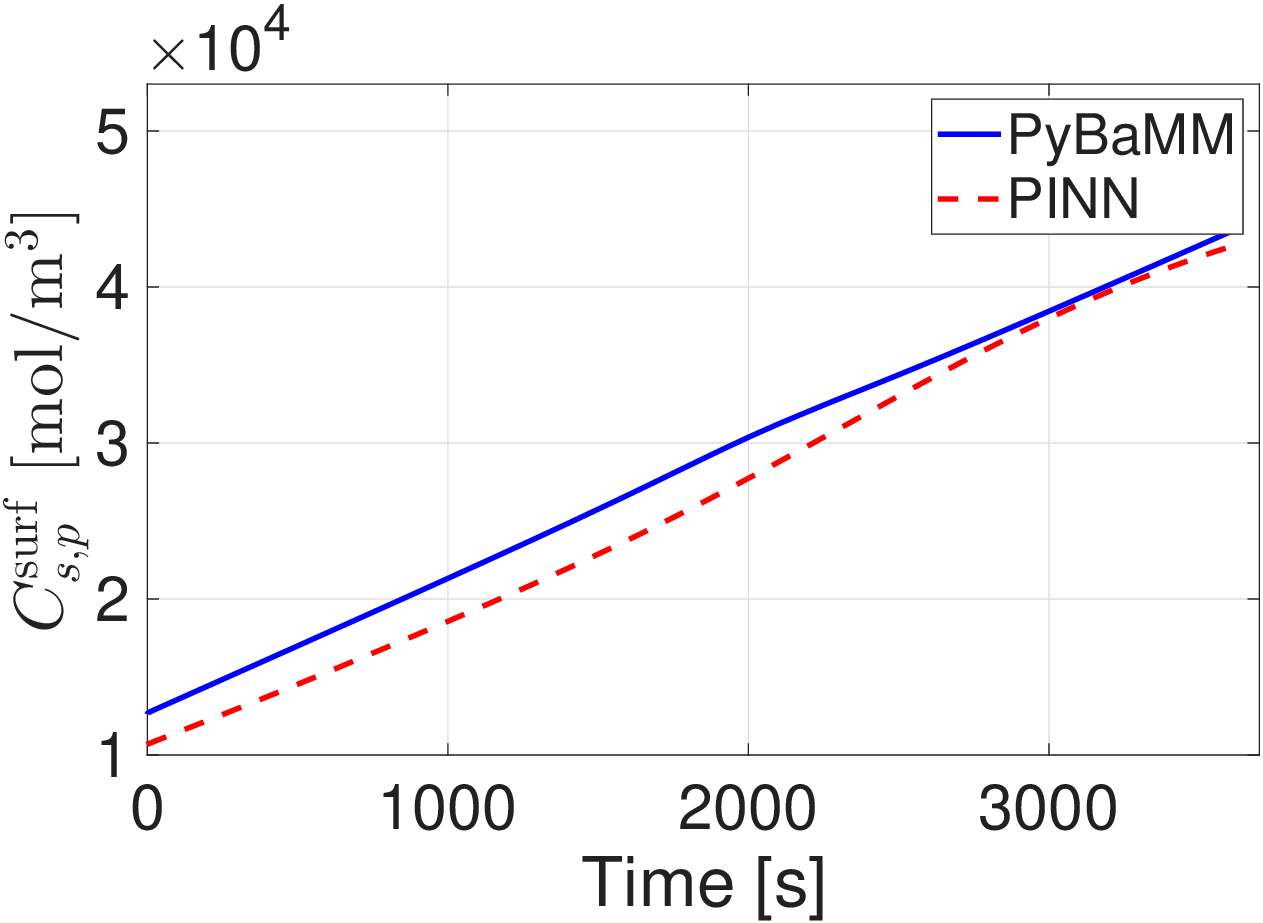}
		\caption{}
		\label{ecker_concentration_positive}
	\end{subfigure}
	\hspace{0.04\textwidth}
	\begin{subfigure}[b]{0.2\textwidth}
		\centering
		\includegraphics[height=3cm,keepaspectratio]{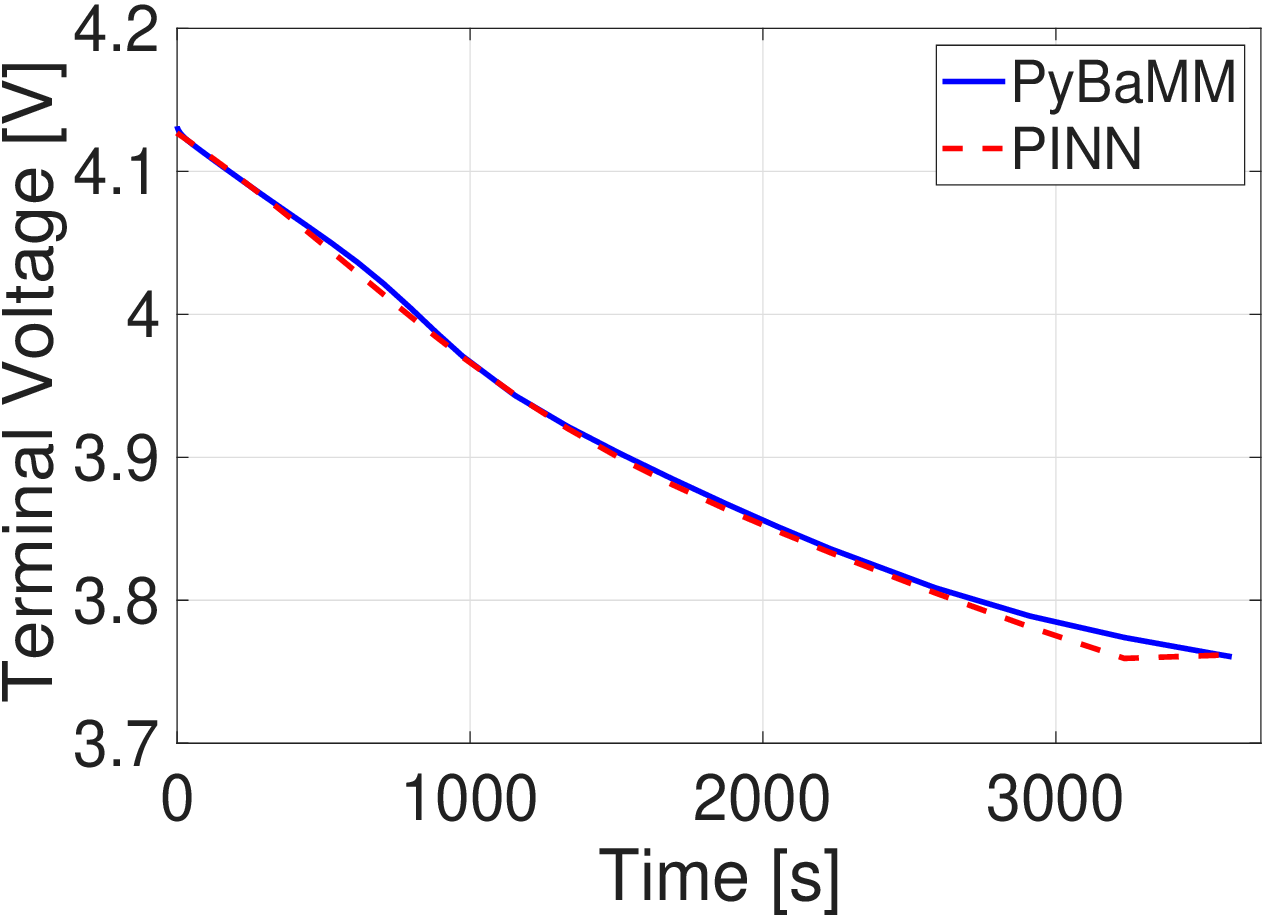}
		\caption{}
		\label{ecker_0.5}
	\end{subfigure}
	\caption{Comparison between the PyBaMM SPMe model and the transfer learning in the SPMe-PINN framework for B2 showing (a) terminal voltage prediction at a 1C discharge rate, (b) negative electrode solid-phase lithium concentration, (c) positive electrode solid-phase lithium concentration, and (d) terminal voltage prediction at a 0.5C discharge rate.}
\end{figure}

To evaluate the transferability of the proposed framework across different cell configurations, we consider the B2 dataset, which is based on a Kokam cell with graphite/NMC chemistry and a pouch cell format. In contrast to the source battery, which is cylindrical, this introduces variations in cell geometry and electrochemical response, making it a suitable test case for transfer learning. In this stage, the pretrained SPMe-PINN model is adapted to the target battery by freezing some layers and fine-tuning the remaining network parameters. The model is evaluated under a 1C discharge condition. The terminal voltage prediction, shown in Fig. \ref{ecker_voltage}, demonstrates strong agreement with the PyBaMM reference solution, although an increase in error is observed relative to the source case due to domain shift.

Additionally, the model is used to estimate internal electrochemical states, specifically the surface lithium concentration at both electrodes. As illustrated in Fig. \ref{ecker_concentration_negative} and Fig. \ref{ecker_concentration_positive}, the predicted concentration profiles closely match the ground truth, confirming that the model retains its ability to capture internal dynamics after transfer.

To further assess the robustness, we evaluate the model under a different operating condition using the same B2 dataset. Specifically, a 0.5C discharge rate is applied for 1 hour. The results, shown in Fig. \ref{ecker_0.5}, indicate that the proposed model maintains a high prediction accuracy for the terminal voltage. 

\subsection{4.3 Cross-Chemistry Transfer}
\begin{figure}[h]
	\centering
	% Top row
	\begin{subfigure}[b]{0.2\textwidth}
		\centering
		\includegraphics[height=3cm,keepaspectratio]{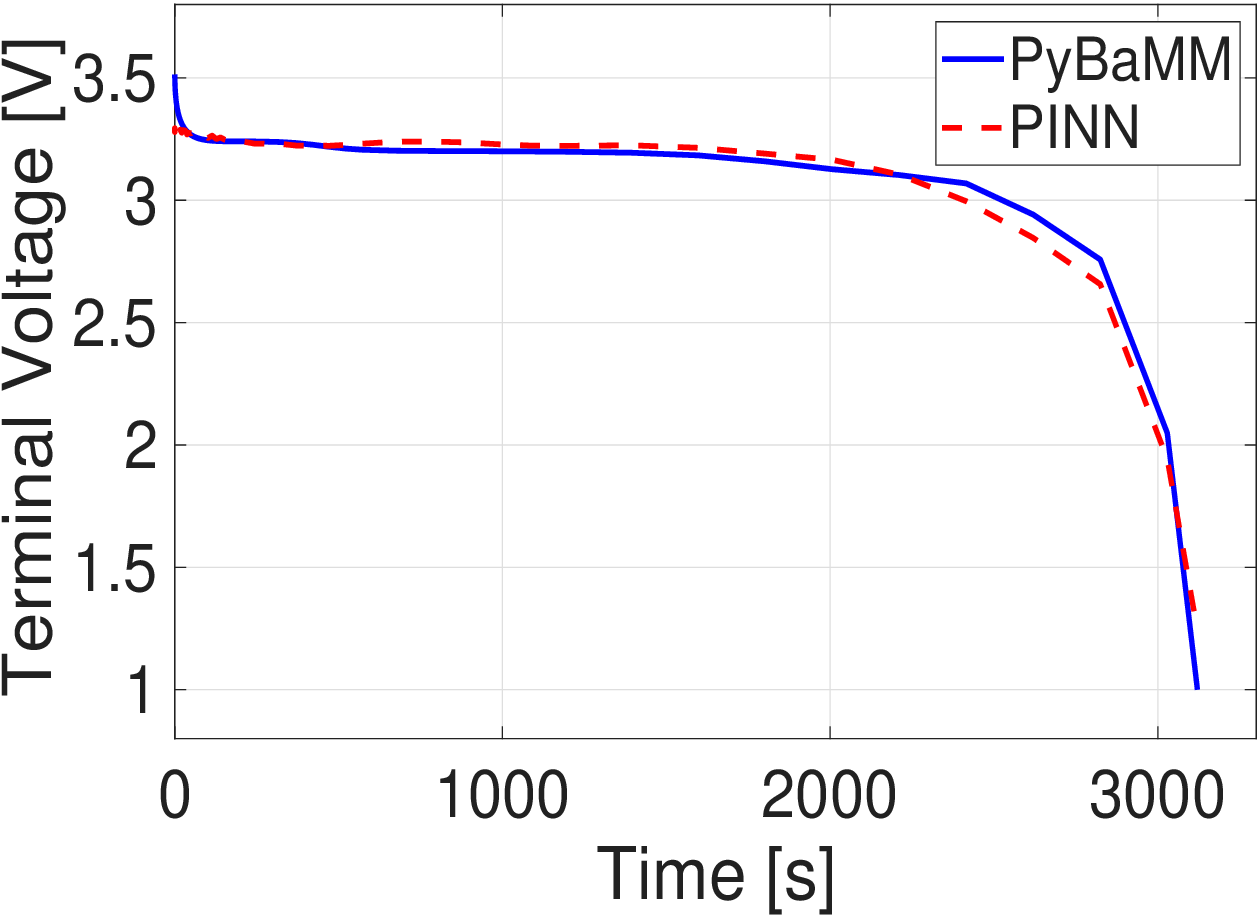}
		\caption{}
		\label{prada_voltage}
	\end{subfigure}
	\hspace{0.04\textwidth}
	\begin{subfigure}[b]{0.2\textwidth}
		\centering
		\includegraphics[height=3cm,keepaspectratio]{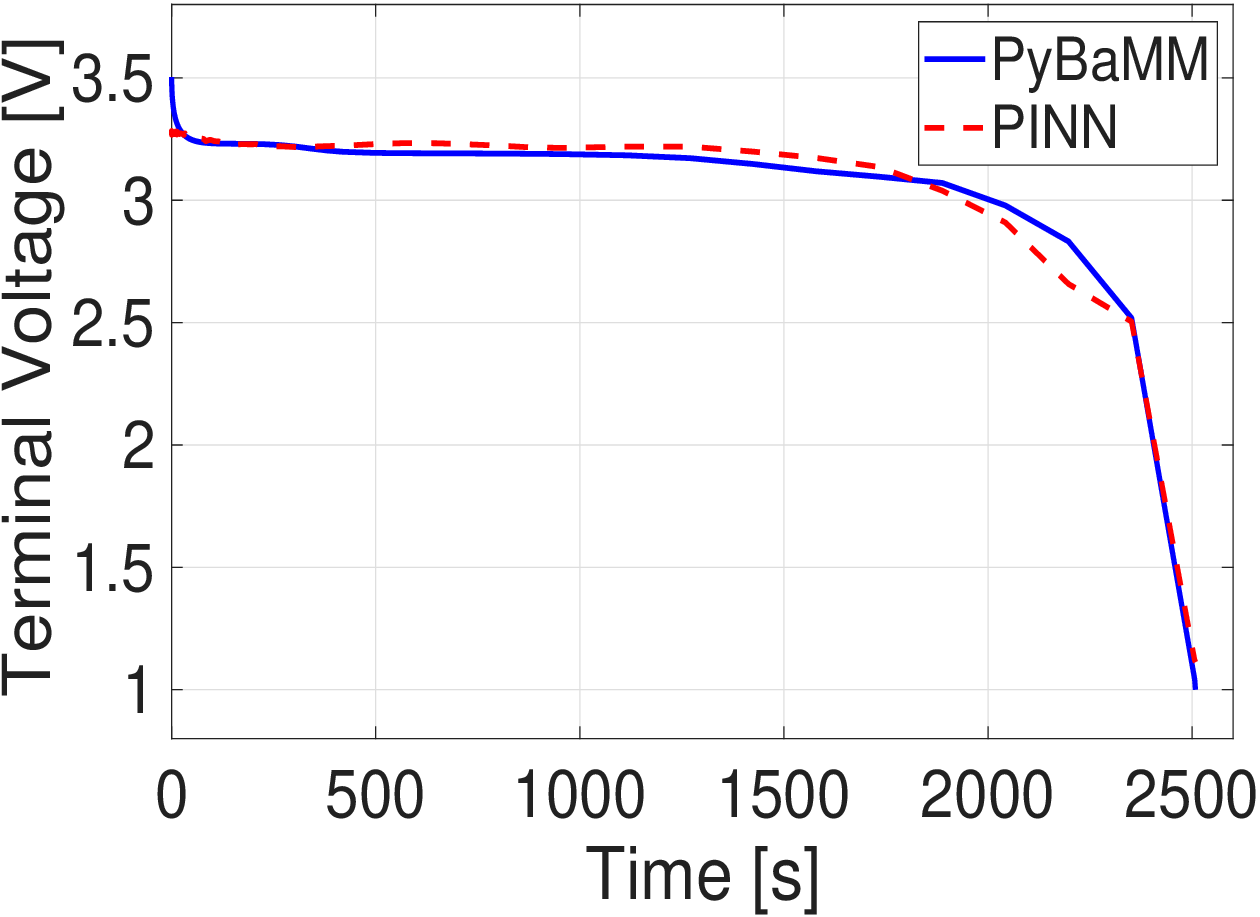}
		\caption{}
		\label{prada_voltage_2}
	\end{subfigure}
	\label{prada_results}
	\caption{Comparison between the PyBaMM SPMe model and the proposed transfer learning SPMe-PINN framework for the B3 showing (a) terminal voltage prediction at 1C discharge rate and (b) terminal voltage prediction at 1.2C discharge rate.}
\end{figure}

To evaluate generalization across different battery chemistries, we consider the B3 dataset, which corresponds to an LFP cell. This represents a more challenging transfer scenario due to the distinct electrochemical characteristics of LFP compared to NMC systems. The same transfer learning procedure is applied, with partial freezing of the pretrained network and fine-tuning on the target dataset. The model is evaluated under both 1C and 1.2C discharge conditions. The predicted terminal voltage are shown in Fig. \ref{prada_voltage} and \ref{prada_voltage_2}. Despite the significant change in chemistry and operating conditions, the model demonstrates strong predictive capability and successfully captures the key features of the LFP voltage response, indicating good generalization performance across different discharge rates.

\subsection{4.4 Parameter Estimation}
To evaluate the parameter identification capability of the proposed framework, the solid-phase diffusivities are modeled as learnable parameters during training. As shown in Table \ref{tab:electrode_comparison}, the predicted diffusivity values exhibit strong agreement with the corresponding ground-truth values, demonstrating the effectiveness of the model in predicting key electrochemical parameters.
\begin{table}[h]
	\centering
	\caption{Comparison between true and predicted electrode diffusivities.}
	\resizebox{0.48\textwidth}{!}{
	\begin{tabular}{l|l|l|l|l}
		\hline
		\multirow{2}{*}{Battery} & \multicolumn{2}{l|}{Negative Electrode} & \multicolumn{2}{l}{Positive Electrode} \\ \cline{2-5}
		& True & Predicted & True & Predicted \\ \hline
		B2 & 8.332e$^{-15}$  & 8.341e$^{-15}$ & 2.981e$^{-13}$ & 2.983e$^{-13}$  \\ \hline
		B3 & 3.000e$^{-15}$  & 3.013e$^{-15}$ & 5.900e$^{-13}$ &  5.896e$^{-13}$ \\ \hline
	\end{tabular}
}
	\label{tab:electrode_comparison}
\end{table}

\section{5. Conclusion}
\label{conclusion}
In this work, an SPMe-PINN framework was pro-posed for Lithium-ion battery modeling and state estimation. This study highlights the potential of physics-informed machine learning for accurate and scalable battery health modeling and battery management applications. Moreover, the proposed transfer learning-PINN method can effectively adapt pretrained electrochemical knowledge across different battery chemistries with reduced training effort. Future work will focus on incorporating battery degradation and temperature-dependent dynamics to improve model realism. Additional studies will investigate the use of real experimental datasets and adaptive parameter estimation to address battery aging and improve model accuracy under varying operating conditions.

%\section{Acknowledgment}
%Q. Lu acknowledges the startup funds from Texas Tech University. S. Al-Wahaibi acknowledges the Distinguished Graduate Student Assistantships from Texas Tech University.

%\input{tex/conclusion}

%-----------------------------------------------------------------------------------------------------------------------------------------------------------
%
%
%-----------------------------------------------------------------------------------------------------------------------------------------------------------

\bibliographystyle{chicago}
\section{References}
%\parindent=0cm \textbf{References}

%\noindent{\bf \textsc{Key references:}}
%\vspace{-15mm}

\def\refname{}
\def\bibsection{}

\bibliography{references_arxiv.bib}%

@article{khan2023design,
  title={Design and optimization of lithium-ion battery as an efficient energy storage device for electric vehicles: A comprehensive review},
  author={Khan, FM Nizam Uddin and Rasul, Mohammad G and Sayem, ASM and Mandal, Nirmal K},
  journal={Journal of Energy Storage},
  volume={71},
  pages={108033},
  year={2023},
  publisher={Elsevier}
}

@article{ekberg2025state,
  title={State estimation and remaining useful life prediction for lithium-ion batteries},
  author={Ekberg, Johan and Sridharan, Naveen Venkatesh and Karim, Ramin and Atta, Khalid Tourkey},
  journal={Cell Reports Physical Science},
  year={2025},
  publisher={Elsevier}
}

@article{doyle1993modeling,
  title={Modeling of galvanostatic charge and discharge of the lithium/polymer/insertion cell},
  author={Doyle, Marc and Fuller, Thomas F and Newman, John},
  journal={Journal of the Electrochemical society},
  volume={140},
  number={6},
  pages={1526--1533},
  year={1993},
  publisher={The Electrochemical Society, Inc.}
}

@article{moura2016battery,
  title={Battery state estimation for a single particle model with electrolyte dynamics},
  author={Moura, Scott J and Argomedo, Federico Bribiesca and Klein, Reinhardt and Mirtabatabaei, Anahita and Krstic, Miroslav},
  journal={IEEE Transactions on Control Systems Technology},
  volume={25},
  number={2},
  pages={453--468},
  year={2016},
  publisher={IEEE}
}

@article{santhanagopalan2006review,
  title={Review of models for predicting the cycling performance of lithium ion batteries},
  author={Santhanagopalan, Shriram and Guo, Qingzhi and Ramadass, Premanand and White, Ralph E},
  journal={Journal of power sources},
  volume={156},
  number={2},
  pages={620--628},
  year={2006},
  publisher={Elsevier}
}

@article{alharbi2025lithium,
  title={Lithium-ion battery state of health degradation prediction using deep learning approaches},
  author={Alharbi, Talal and Umair, Muhammad and Alharbi, Abdulelah},
  journal={IEEE Access},
  volume={13},
  pages={13464--13481},
  year={2025},
  publisher={IEEE}
}

@article{liu2022review,
  title={A review of lithium-ion battery state of charge estimation based on deep learning: Directions for improvement and future trends},
  author={Liu, Yuefeng and He, Yingjie and Bian, Haodong and Guo, Wei and Zhang, Xiaoyan},
  journal={Journal of Energy Storage},
  volume={52},
  pages={104664},
  year={2022},
  publisher={Elsevier}
}

@article{raissi2019physics,
  title={Physics-informed neural networks: A deep learning framework for solving forward and inverse problems involving nonlinear partial differential equations},
  author={Raissi, Maziar and Perdikaris, Paris and Karniadakis, George E},
  journal={Journal of Computational physics},
  volume={378},
  pages={686--707},
  year={2019},
  publisher={Elsevier}
}

@article{singh2023hybrid,
  title={Hybrid modeling of lithium-ion battery: Physics-informed neural network for battery state estimation},
  author={Singh, Soumya and Ebongue, Yvonne Eboumbou and Rezaei, Shahed and Birke, Kai Peter},
  journal={Batteries},
  volume={9},
  number={6},
  pages={301},
  year={2023},
  publisher={MDPI}
}

@article{nascimento2021hybrid,
  title={Hybrid physics-informed neural networks for lithium-ion battery modeling and prognosis},
  author={Nascimento, Renato G and Corbetta, Matteo and Kulkarni, Chetan S and Viana, Felipe AC},
  journal={Journal of Power Sources},
  volume={513},
  pages={230526},
  year={2021},
  publisher={Elsevier}
}

@article{mendez2024physics,
  title={Physics-informed neural networks for modeling li-ion batteries: Solving the single particle model without labeled data},
  author={M{\'e}ndez-Corbacho, Francisco J and Larrarte-Lizarralde, Be{\~n}at and Parra, Rub{\'e}n and Larrain, Javier and del Olmo, Diego and Grande, Hans-J{\"u}rgen and Ayerbe, Elixabete},
  journal={Journal of The Electrochemical Society},
  volume={171},
  number={11},
  pages={110534},
  year={2024},
  publisher={IOP Publishing}
}

@article{xue2023enhanced,
  title={An enhanced single-particle model using a physics-informed neural network considering electrolyte dynamics for lithium-ion batteries},
  author={Xue, Chenyu and Jiang, Bo and Zhu, Jiangong and Wei, Xuezhe and Dai, Haifeng},
  journal={Batteries},
  volume={9},
  number={10},
  pages={511},
  year={2023},
  publisher={MDPI}
}

@article{mehta2021improved,
  title={An improved single-particle model with electrolyte dynamics for high current applications of lithium-ion cells},
  author={Mehta, Rohit and Gupta, Amit},
  journal={Electrochimica Acta},
  volume={389},
  pages={138623},
  year={2021},
  publisher={Elsevier}
}

@article{chen2020development,
  title={Development of experimental techniques for parameterization of multi-scale lithium-ion battery models},
  author={Chen, Chang-Hui and Brosa Planella, Ferran and O’regan, Kieran and Gastol, Dominika and Widanage, W Dhammika and Kendrick, Emma},
  journal={Journal of The Electrochemical Society},
  volume={167},
  number={8},
  pages={080534},
  year={2020},
  publisher={IOP Publishing}
}

@article{ecker2015parameterization,
  title={Parameterization of a physico-chemical model of a lithium-ion battery: I. Determination of parameters},
  author={Ecker, Madeleine and Tran, Thi Kim Dung and Dechent, Philipp and K{\"a}bitz, Stefan and Warnecke, Alexander and Sauer, Dirk Uwe},
  journal={Journal of The Electrochemical Society},
  volume={162},
  number={9},
  pages={A1836--A1848},
  year={2015},
  publisher={The Electrochemical Society}
}

@article{prada2013simplified,
  title={A simplified electrochemical and thermal aging model of LiFePO4-graphite Li-ion batteries: power and capacity fade simulations},
  author={Prada, Eric and Di Domenico, D and Creff, Y and Bernard, J and Sauvant-Moynot, Val{\'e}rie and Huet, Fran{\c{c}}ois},
  journal={Journal of The Electrochemical Society},
  volume={160},
  number={4},
  pages={A616--A628},
  year={2013},
  publisher={The Electrochemical Society}
}

@article{marquis2019asymptotic,
  title={An asymptotic derivation of a single particle model with electrolyte},
  author={Marquis, Scott G and Sulzer, Valentin and Timms, Robert and Please, Colin P and Chapman, S Jon},
  journal={Journal of The Electrochemical Society},
  volume={166},
  number={15},
  pages={A3693--A3706},
  year={2019},
  publisher={The Electrochemical Society}
}

%\bibliographystyle{unsrt}

%\end{spacing}
%\begin{thebibliography}{}
%
%
%\bibitem[\protect\citeauthoryear{Ydstie}{Ydstie}{2004}]{ydstie2004distributed}
%Ydstie, B.E. (2004).
%\newblock Distributed decision making in complex organizations: the adaptive
%  enterprise.
%\newblock {\em Computers \& chemical engineering\/}~{\em 29\/}(1), 11--27.
%
%
%
%\end{thebibliography}

%\end{multicols}

\end{document}